\documentclass{article}
\usepackage{spconf,amsmath,graphicx,comment,hyperref}
\usepackage{xcolor}

\title{SALYPATH: A DEEP-BASED ARCHITECTURE FOR VISUAL ATTENTION PREDICTION}
\address{Author Affiliation(s)}

\name{Mohamed A. KERKOURI $^{1}$\qquad  Marouane TLIBA$^{2}$ \qquad Aladine CHETOUANI $^{1}$ \qquad Rachid HARBA$^{1}$
\thanks{Funded by the TIC-ART project, Regional fund (Region Centre-Val de Loire). \newline \quad  Code can be accessed at: \url{https://github.com/kmamine/SALYPATH}. }}

\address{$^{1}$Laboratoire PRISME, Université d'Orléans, Orléans, France\\
$^{2}$Institut National Des Télécommunications et TIC, Oran, Algeria\\}

\begin{document}
\ninept
\maketitle 
%

\begin{abstract}
Human vision is naturally more attracted by some regions within their field of view than others. This intrinsic selectivity mechanism, so-called visual attention, is influenced by both high- and low-level factors; such as the global environment (illumination, background texture, etc.), stimulus  characteristics (color, intensity, orientation, etc.), and some prior visual information. Visual attention is useful for many computer vision applications such as image compression, recognition, and captioning. In this paper, we propose an end-to-end deep-based method, so-called SALYPATH (SALiencY and scanPATH), that efficiently predicts the scanpath of an image through features of a saliency model. The idea is predict the scanpath by exploiting the capacity of a deep-based model to predict the saliency. The proposed method was evaluated through 2 well-known datasets. The results obtained showed the relevance of the proposed framework comparing to state-of-the-art models.






\end{abstract}

\begin{keywords}
Visual attention, eye movement, saliency, scanpath prediction   
\end{keywords}
\vspace{-3mm}

\section{Introduction}


Human vision is naturally more attracted by some regions within their field of view than others. This natural selectivity mechanism, so-called visual attention, is influenced by both high- and low-level factors; such as the global environment (illumination, background texture, etc.), stimulus  characteristics (color, intensity, orientation, etc.), and some prior visual information \cite{featureIntegration}. It gives the Human Visual System (HVS) an astonishing efficiency for detection and recognition through rapid eye movements called saccades. Predicting visual attention became a staple to improve many image processing and computer vision applications such as indoor localization \cite{VCIP14}, image quality \cite{SPIC20Chetouani,ICIP18Ilyass,PR20Ilyass, EUSIPCO18Chetouani}, image watermarking \cite{info19Hamidi}, image compression, recognition, and captioning. Visual attention is usually depicted using 2D heat maps, often called saliency maps, representing the most spatial-attractive regions in a given stimulus. The sequential representation of the points follows during the exploration of the image, also called scanpath, is used to derive saliency maps.

Researchers put a lot of efforts in predicting such saliency heat maps, starting with the seminal work of Koch and Ulman \cite{KochUllman}, later followed by Itti et al. \cite{Itti} where multi-scale low level features were used. In \cite{GBVS}, the saliency is predicted based on the graph theory where Markov chains is defined over different input  maps. Several other interesting saliency models have been proposed based on heuristic approaches and low level features  \cite{spectral1} \cite{spectral2}.     


With the recent success of deep learning, many deep-based saliency models have been developed. Pan et al. \cite{PanDSN} proposed one of the first Convolutional Neural Network (CNN) models where a deep and shallow networks were used. In \cite{salgan}, the authors proposed a deep convolutional Generative Adversarial Network (GAN) based saliency model with adversarial training. In \cite{deepgaze1}, the authors introduced a method called DeepGaze I where they trained an AlexNet \cite{deepgaze1ref} based network on MIT dataset. The authors later introduced a new version, called DeepGazeII \cite{deepgaze2}, where a VGG-19 network \cite{VGG} was exploited. In \cite{MLNet}, the authors proposed a more complex architecture where multi level features extracted from a VGG \cite{VGG} network were employed. In \cite{SAM}, the authors designed a method to predict the saliency by incorporating an attention mechanism based on the combination of Long Short Term Memory (LSTM) and convolutional networks. In \cite{unisal}, the authors proposed a unified model for saliency prediction, called Unisal, that predicts the saliency for images and videos. 
\vspace{-3mm}


Contrary to the large number of saliency models proposed in the literature, the studies dedicated to the prediction of scanpath are less extensive. In \cite{lemeur}, the authors proposed a model where visual scanpaths are inferred stochastically from saliency maps as well as saccadic amplitudes and orientation biases derived from several datasets. In \cite{saltinet}, the authors proposed a deep model, called Saltinet, that samples the saccades from a predicted static saliency volume generated by a CNN encoder-decoder network. The authors later proposed a saliency model, called PathGan \cite{pathgan}, that predicts scanpath by using LSTM layers and a VGG network with adversarial training. The underlying idea of using LSTM was to predict the current fixation point according to the previous ones to increase the sequential dependence between fixations. In \cite{G-Eymol}, the authors presented a model of scanpath as a dynamic process simulating the laws of gravitational mechanics. The gaze is considered as traveling mass point in the image space, and salient regions as gravitational fields affecting the mass speed and trajectory. In \cite{DCSM}, the authors proposed a Deep Convolutional Saccadic Model (DCSM) where the fixation points are predicted from foviated saliency maps and temporal duration while modeling the inhibition of return.

\begin{figure*}[ht!]
  \centering
  \centerline{\includegraphics[height=50mm]{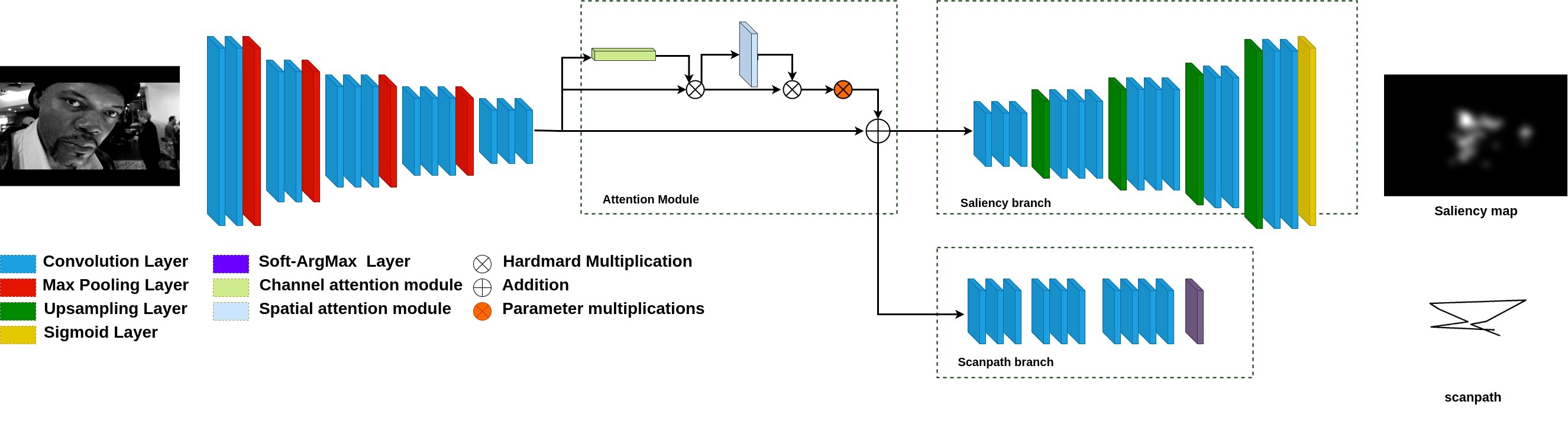}}
\caption{\textbf{SALYPATH:} SALiencY and scanPATH prediction model}
\label{fig:arch}
\end{figure*}
\vspace{-3mm}

In this study, we propose an end-to-end deep-based method, called SALYPATH (SALiencY and scanPATH), that efficiently predicts the scanpath of images by leveraging features from a trained saliency model. The idea is to predict the scanpath of a given image by exploiting the capacity of a deep-based model to predict the saliency. Inspired by SalGan \cite{salgan}, we first construct a saliency model that integrates an attention module. Feature maps are then extracted from the designed saliency model and fed as input to a second CNN model that aims to predict the scanpath of the image.



The rest  of  the  paper  is organized as follows: Section 2 describes the proposed method in details. We then discuss the results obtained in Section 3. Finally, we provide some conclusions in Section 4.

\section{Proposed Method}

As illustrated by Fig. \ref{fig:arch}, we propose a novel fully convolutional neural network architecture for predicting saliency and scanpath of static natural images. Our model is composed of a VGG-based encoder as well as two predictor networks to predict the corresponding saliency and scanpath, respectively. An attention module is also used to adaptively refine the features extracted from the encoder part. Each part of the proposed model is described in this section.


\subsection{Saliency Prediction}

In order to predict the scanpath of a given image, we design a deep-based saliency model. The idea is to predict the scanpath of a given image by exploiting high level features of a deep-based saliency model. The proposed model is constituted by a VGG-based encoder-decoder similar to the generator part of SalGan \cite{salgan}. The encoder part is composed of (3x3) convolutional layers and (2x2) max-pooling layers, while the decoder part is composed of (3x3) convolutions and (2x2) up-sampling layers with a (1x1) final convolutional layer activated by a Sigmoid function. The encoder aims to generate high dimensional feature maps from the input image by aggregating them on multiple levels, while decoder aims to merge the obtained feature maps to a single image that represents the saliency map.

\subsection{Scanpath Prediction}

From the model's bottle-neck, we extract the high level representational features to predict the scanpath through a fully convolutional network. It is composed of 10 convolutional layers while gradually decreasing the depth of the feature maps to 8 channels, corresponding to the length of the predicted scanpath. The latter has been fixed according to a statistical analysis of the lengths of scanpaths provided by the dataset used to train our model (see Section \ref{DATASET}). The central tendency indicates that a length of 8 fixation points is appropriate. The output of our fully convolutional network is then fed to a \textit{Soft-ArgMax} (SAM) \cite{Soft-ArgMax} function which returns the coordinates of the highest activation point for each feature map as follows:

\begin{equation}
  SAM(x) =  \sum_{i=0}^{W} \sum_{j=0}^{H} \frac{e^{\beta x_{i,j}}}{\sum_{i^\prime =0 }^{W} \sum_{j^\prime = 0 }^{H} e^{\beta x_{i^\prime ,j^\prime }}}({\frac{i}{W},\frac{j}{H}})^T
  \label{eq:SAM}
\end{equation}
where ${i,j,i^\prime ,j^\prime}$ iterate over pixel coordinates and ${H,W}$ represent the height and width of the feature map, respectively. $x$ is the input feature map and $\beta$ is a parameter adjusting the distribution of the softmax output. 

It is worth noting that unlike the discrete \textit{ArgMax}, the \textit{Soft-ArgMax} function is differentiable. It also allows a sub-pixel regression, thus providing a better prediction even with smaller feature maps passed on the scanpath prediction model.  

\vspace{-3mm}

\subsection{Attention Module}


In other fields of computer vision, attention is an approach for attending to different parts of an input vector to capture a global representation of features. It has been proven that adding such modules allows to refine intermediate features and thus improve the global performance \cite{ChenAtt} \cite{xu2015show}. Thanks to this specificity, an attention module has been here incorporated to improve the feature representation used to predict both saliency and scanpath. More precisely, we employed the Convolutional Block Attention Module (CBAM) \cite{CBAM} which is composed of 2 blocks as shown in the attention module box in Fig.\ref{fig:arch}. The channel attention block exploits the interrelations between feature maps and outputs a channel-wise weighted feature vector emphasizing the more important features in relation to the task. While the spatial attention block calculates the importance of features according to their spatial position emphasizing the most important location in each high-level feature. The attention module is used in parallel to the stream between encoder and decoder at the bottle-neck. The output of the module is represented by the following equation:  

\begin{equation}
  z = X \otimes Att_S(X \otimes Att_{Ch}(X))
  \label{eq:cbam_out}
\end{equation}
where \textit{z} is the output of the module, \textit{$Att_S$} and \textit{$Att_{Ch}$} are the spatial attention and the channel attention blocks, respectively. $X$ is the input feature maps and $\otimes$ represents the element-wise multiplication.

The input $X^\prime$ of the decoder and the fully convolutional network branches is then computed as follows:
\begin{equation}
  X^\prime = X + z \otimes \gamma 
  \label{eq:input_decode}
\end{equation}
where $\gamma$ is a learnable parameter.

\begin{table*}[ht!]
\begin{center}
\scalebox{0.8}{
\begin{tabular}{ c  c  c  c  c  c  c  c c}
\hline
\textbf{Model} & \textbf{Auc Judd $\uparrow$} & \textbf{Auc Borji$\uparrow$} & \textbf{NSS$\uparrow$}& \textbf{CC$\uparrow$}& \textbf{SIM$\uparrow$} & \textbf{KLD $\downarrow$}  \\ 

\hline
 Salgan\cite{salgan} & 0.8662  & \textbf{0.8443}  & 1.9460  &  0.5836 & 0.4908  & 1.0470         \\ 
 \hline
 Salicon\cite{opensalicon} &  0.8630 & 0.8135   & 2.1240 &  0.6013  &  0.4923     & 0.9051      \\ 
 \hline
 MLNet\cite{MLNet} & 0.8509 & 0.7714 & \textbf{2.1678} & 0.5787 & 0.4815 & 1.3083 \\
 \hline
SALYPATH (Our method)  & \textbf{0.8745} & 0.8290 & 2.1152 & \textbf{0.6182} &  \textbf{0.4982}   & \textbf{0.8750}       \\ 
\hline
 %

\end{tabular}}
\caption{\label{tab:sal_mit_results}Results of saliency prediction on MIT1003.}
\end{center}
\end{table*}
\vspace{-3mm}

\subsection{Training}

Two different loss functions have been used to train each branch of our model. The saliency branch was trained using the following loss function $L_1$:
\begin{equation}
\begin{split}
    \textit{$L_1$} = 0.6 \times  KLdiv(y,\hat{y} ) + 0.3 \times  MSE(y,\hat{y} ) \\ - 0.1 \times NSS(y,\hat{y})
     \label{eq:SAlL}
\end{split}
\end{equation}
where $KLdiv$ is the Kullback-Leibler Divergence, $MSE$ is the Mean Squared Error and $NSS$ is Normalized Scanpath Saliency. $y$ and $\hat{y}$ are the predicted and the ground truth saliency maps, respectively. 

Each term of the designed loss function has its own impact on the convergence of our model \cite{Losses}. Indeed, $KLdiv$ function aims to compare the distributions of the output and the corresponding ground truth, while $MSE$ function regularizes the loss by comparing the predicted saliency map and its corresponding ground truth on pixel level \cite{LossesSal}. We also introduced the NSS value which is usually used as metric to evaluate saliency prediction \cite{NSS}. It allows to capture several properties that are specific to saliency maps. This branch was trained with a learning rate of $10^{-7}$ and a step LR scheduler with a multiplicative factor of 0.9 per epoch.

The scanpath branch was trained using only the $MSE$ as loss function $L_2$ (see eq. \ref{eq:SPL}). It was used since scanpath prediction are characterized by the locations of their fixation points and thus their prediction can be seen as a regression problem. It worth noting that the scanpaths are predicted through features of the designed saliency model. To better predict the scanpath, we here focused more on improving the intermediate representational space, optimized through the more complex saliency loss function $L_1$ (see eq. \ref{eq:SAlL}). This branch was trained with a learning rate of $10^{-5}$ and a step LR scheduler with a multiplicative factor of 0.9 per epoch.




\begin{equation}
    L_2 = \frac{1}{N} \sum_i (p - \hat{p} )^2
    \label{eq:SPL}
\end{equation}
where $p$ is the predicted scanpath and $\hat{p}$ is the ground truth scanpath, while $N$ is the number of the fixation points.
\vspace{-3mm}

\section{Experimental Results}
In this section, we evaluate the capacity of our method to predict the scanpath of a given image. After presenting the datasets used, the saliency prediction branch as well as the scanpath prediction branch are evaluated. Both are also compared to a set of representative state-of-the-art methods.


\subsection{Datasets}
\label{DATASET}
In order to evaluate of our method, two widely used datasets have been employed: Salicon \cite{Salicon} and MIT1003 \cite{MIT1003}. Salicon dataset is the largest natural image saliency dataset and it was built for the Salicon challenge. More precisely, we used a subset of 15000 images from the dataset for training (i.e. 9000), validation (i.e. 1000) and testing (i.e. 5000). MIT1003 dataset is one of the most used natural image saliency datasets and it was employed during the MIT300 challenge\cite{MIT300}. It is composed of 1003 natural images with their corresponding saliency maps and scanpaths. The whole dataset has been considered in this study for cross-dataset testing.



\subsection{Saliency Prediction}

In this section, we evaluate the saliency prediction branch using common metrics \cite{salMetrics_Bylinskii}: Area Under Curve Judd ($Auc\_Judd$), $Auc\_Borji$, NSS, Correlation coefficient ($CC$), Similarity ($SIM$) and kullback Leibler Divergence ($KLD$). Table \ref{tab:sal_mit_results} shows the results obtained on the MIT1003 dataset. The results obtained are also compared to some state-of-the-art saliency models (i.e. Salgan \cite{salgan}, Salicon \cite{Salicon} and MLNet \cite{MLNet}). As can be seen, our model achieves the best results compared to the other models in 4 different metrics including $Auc\_Judd$, $CC$, $SIM$, and $KLD$. In particular, we obtain a significant improvement in terms of $KLD$ compared to Salgan and MLNet.
We reach close second place to SalGan in terms of $Auc\_Borji$ metric, while the obtained $NSS$ values are close to Salicon and MLNet.


\subsection{Scanpath Prediction}

\begin{table*}[h]
\begin{center}
\scalebox{0.8}{
\begin{tabular}{ c  c  c  c  c  c  c c}
\hline
\textbf{Model} & \textbf{MM Shape } & \textbf{MM Dir}   & \textbf{MM Len} & \textbf{MM Pos}& \textbf{MM Mean}& \textbf{NSS} & \textbf{Congruency}  \\ 
\hline
 PathGan\cite{pathgan} & 0.9608  &  0.5698   &   \textbf{0.9530} &     0.8172  &    0.8252 & -0.2904  &  0.0825        \\ 
 \hline
 Le Meur\cite{lemeur} & 0.9505  & 0.6231   &  0.9488 & 0.8605  &   0.8457  &  \textbf{0.8780} &   \textbf{0.4784}        \\ 
 \hline
 G-Eymol\cite{G-Eymol} & 0.9338 & 0.6271 & 0.9521 & \textbf{0.8967} &  0,8524 & 0.8727 & 0.3449\\
 
 \hline
 SALYPATH (Our method) & \textbf{0.9659}  & \textbf{0.6275} & 0.9521 & 0.8965  &   \textbf{0,8605} &   0.3472 &  0.4572     \\ 
 \hline
\end{tabular}}
\caption{\label{tab:salicon_results}Results of scanpath prediction on Salicon.}
\end{center}
\end{table*}

\begin{table*}[htp!]
\begin{center}
\scalebox{0.8}{
\begin{tabular}{ c  c  c  c  c  c  c c}
\hline
\textbf{Model} & \textbf{MM Shape } & \textbf{MM Dir}   & \textbf{MM Len} & \textbf{MM Pos}& \textbf{MM Mean}& \textbf{NSS} & \textbf{Congruency}  \\ 
\hline
 PathGan\cite{pathgan} & 0.9237  & 0.5630   & 0.8929   &  0.8124    &   0.7561  & -0.2750  &  0.0209        \\ 
 \hline
 DCSM (VGG)\cite{DCSM}  & 0.8720 & 0.6420 & 0.8730 & 0.8160 & 0,8007 & - & - \\ 
 \hline
 DCSM (ResNet)\cite{DCSM} & 0.8780 & 0.5890 & 0.8580 & \textbf{0.8220} &  0,7868 & - & -\\ 
 \hline
 Le Meur\cite{lemeur} & 0.9241  &  0.6378  & \textbf{0.9171}  &  0.7749 &  0,8135  & 0.8508   &  \textbf{0.1974}    \\ 
 \hline
 G-Eymol\cite{G-Eymol} & 0.8885 & 0.5954 & 0.8580 & 0.7800 &  0,7805 & \textbf{0.8700} & 0.1105\\
 
 \hline
 SALYPATH (Our method) & \textbf{0.9363}  & \textbf{0.6507} & 0.9046 & 0.7983  &   \textbf{0,8225} &  0.1595  &   0.0916    \\ 
 \hline

\end{tabular}}
\caption{\label{tab:mit1003_results}Results of scanpath prediction on MIT1003.}
\end{center}
\end{table*}

\begin{figure*}[htp!]
\center
\scalebox{0.7}{
\begin{tabular}{ccccc}
\includegraphics[height = 20mm,width = 30mm]{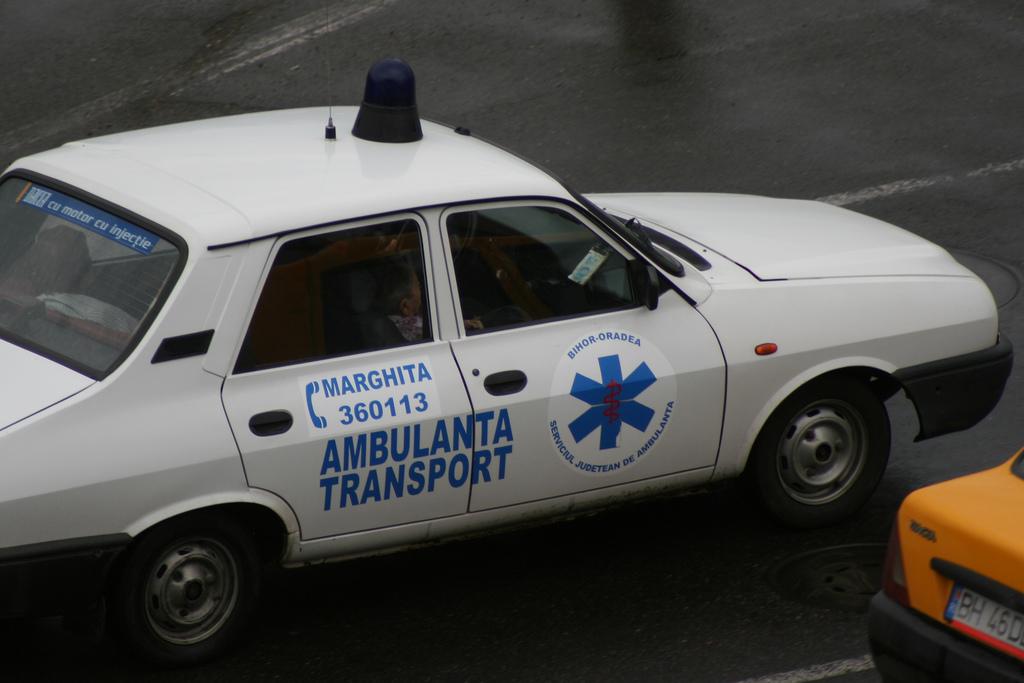} &
\includegraphics[height = 20mm,width = 30mm]{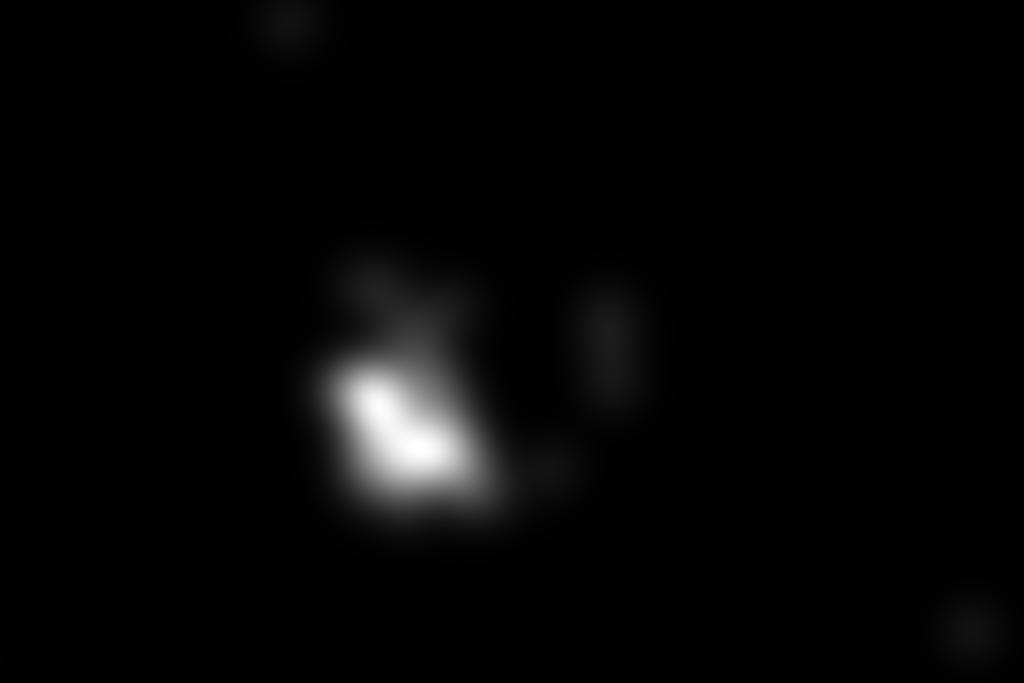} &
\includegraphics[height = 20mm,width = 30mm]{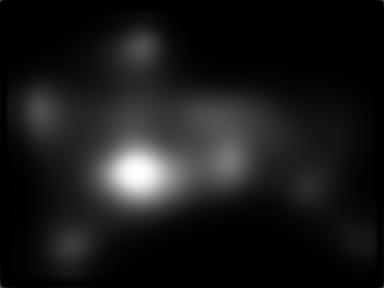} &
\includegraphics[height = 20mm,width = 30mm]{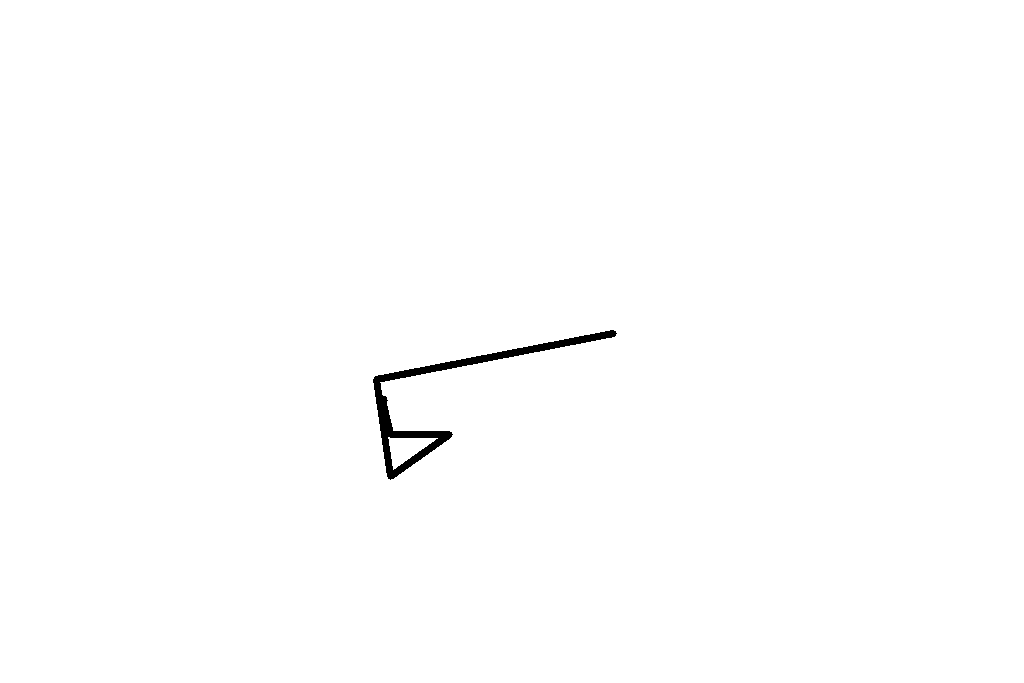}&
\includegraphics[height = 20mm,width = 30mm]{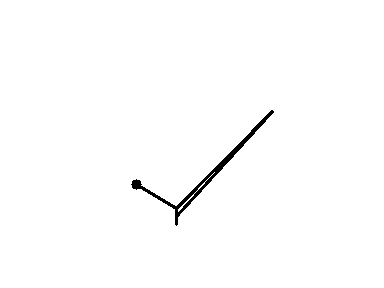} \\

\includegraphics[height = 20mm,width = 30mm]{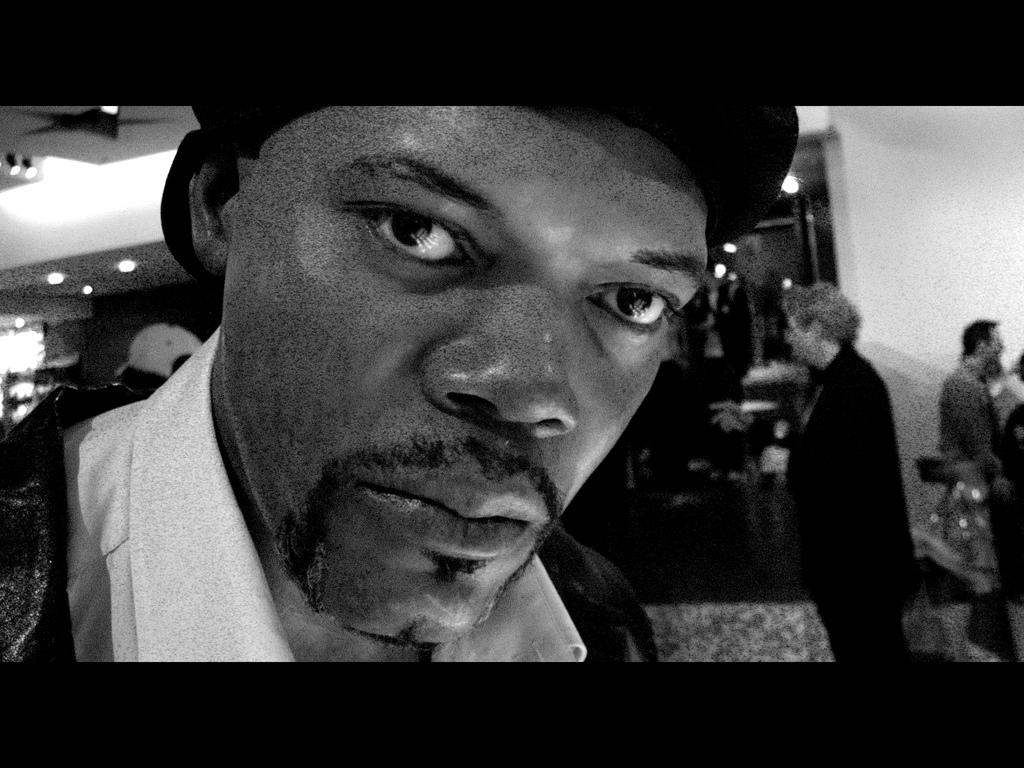} &
\includegraphics[height = 20mm,width = 30mm]{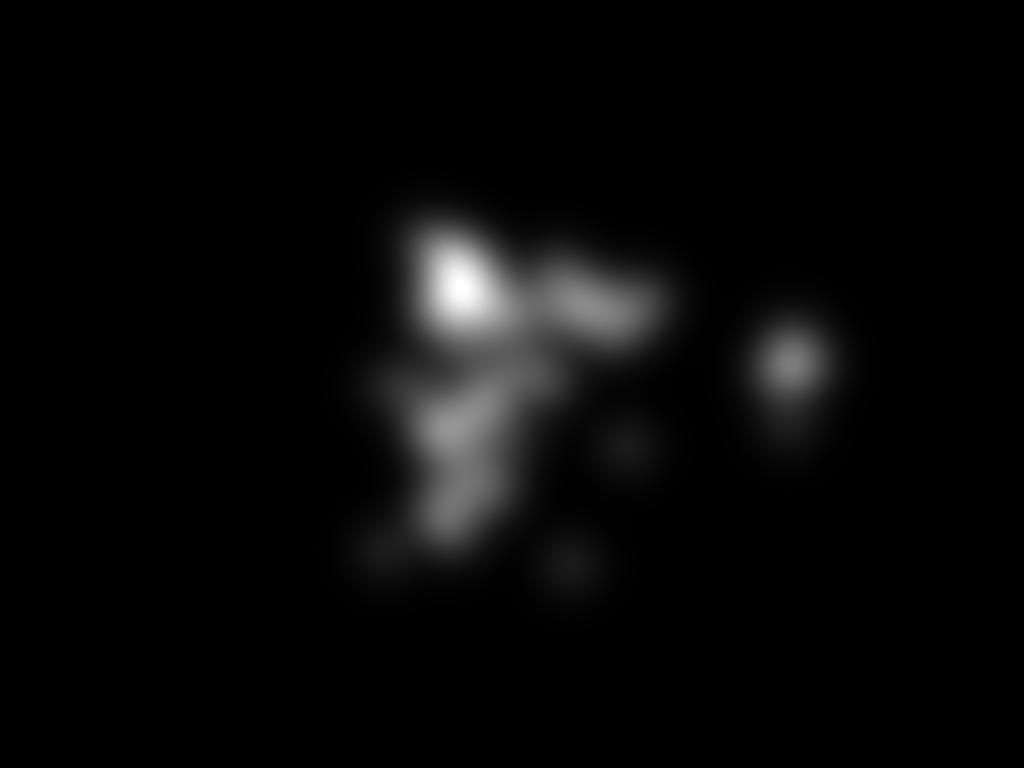} &
\includegraphics[height = 20mm,width = 30mm]{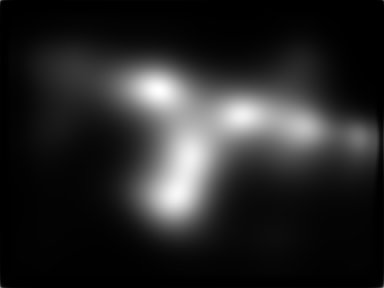} &
\includegraphics[height = 20mm,width = 30mm]{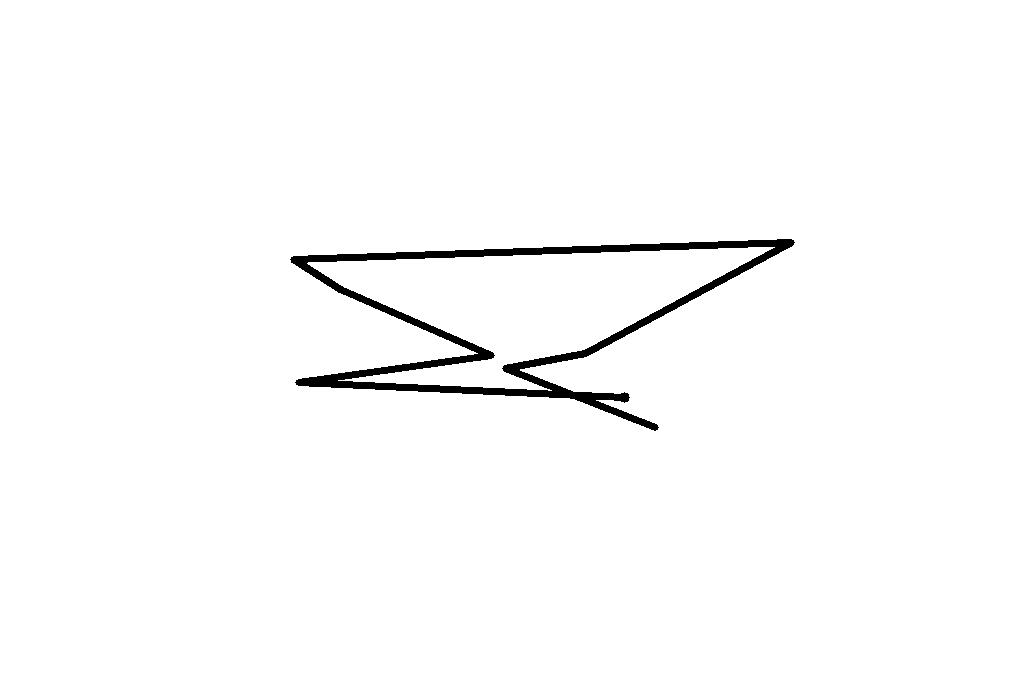}&
\includegraphics[height = 20mm,width = 30mm]{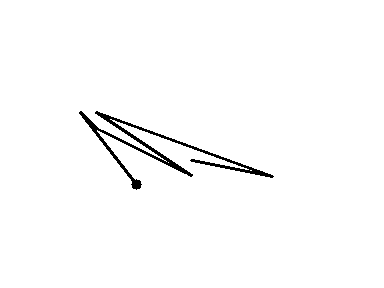}\\

Stimuli &
Ground truth saliency map &
Predicted saliency map &
Ground truth scanpath &
Predicted scanpath \\

\end{tabular}}
\caption{\label{fig:mit1003_viz}Results of scanpath and saliency maps on MIT1003.}
\end{figure*}
\vspace{-3mm}


In order to evaluate the performance of the scanpath prediction branch, three metrics have been used: 
\begin{itemize}
    \item MultiMatch (MM) \cite{Multimatch}: It compares the scanpaths through multiple criteria scores  (i.e. shape of the vectors, the difference in direction and angles between saccades, the length of the saccades, the position of the fixation points and duration between fixation points). In this study, only the first four criteria have been considered since the temporal aspect has not been yet integrated.
    \item NSS \cite{NSS}: It compares a fixation map generated from the predicted fixation points with the ground truth saliency map.
    \item Congruency \cite{congruency}: It measures the percentage of predicted fixation points within a thresholded saliency map.
\end{itemize}

Table \ref{tab:salicon_results} shows the results obtained on Salicon dataset. Our results are also compared to a set of representative state-of-the-art methods, including handcrafted-based models (i.e. Le Meur\cite{lemeur} and G-Eymol \cite{G-Eymol}) and deep learning-based models (i.e. PathGan \cite{pathgan}, DCSM-VGG and DCSM-ResNet \cite{DCSM}). As can be seen, our model achieves the highest values for the shape and direction criteria of the MultiMatch metric. It obtains a very close score to PathGan and G-Eymol on the length and the position criteria, respectively. However, our model obtains the state-of-art results on the mean score. For NSS, our model achieves the third place behind Le Meur and G-Eymol. However, this result can be justified by the fact that models like Le Meur use a predicted saliency map to predict fixation points and thus the fixation points are generally within the salient regions. For the Congruency, our model obtains the second best result close to Le Meur.

Table \ref{tab:mit1003_results} shows the results obtained on MIT1003 dataset which is used as a neutral comparison dataset (i.e. we did not fine-tune our model on it). For MultiMatch metrics, our model scores the highest on the shape and direction criteria, while it achieves a very close second place on the length criterion just behind Le Meur. For the position criterion, our model achieves a better score than Le Meur and G-Eymol, and reasonably close score to PathGan and DSCM-VGG. We still obtain the best overall score for the mean MultiMatch metric. For NSS and Congurency, our method obtains a lower score than Le Meur and G-Eymol but still outperforms PathGan. 

In Fig \ref{fig:mit1003_viz}, we present scanpaths and saliency maps generated by our model as well as their corresponding ground truth scanpaths and saliency maps for two natural images of the MIT1003 dataset. As can be seen, the visualization demonstrates the efficiency of our model for both branches. 
\vspace{-3mm}


\section{Conclusion}

In this paper, we proposed a novel fully convolutional neural network architecture for predicting saliency and scanpaths of natural images. Our model is composed of an encoder-decoder to predict the saliency and a second branch from which the scanpath is predicted. The latter takes advantage of the features provided at the bottle-neck of the designed saliency model. An attention module was also used to refine the image encoded feature space, improving thus the saliency and scanpath prediction.
The proposed model outperformed a good representative set of state-of-the-art models for saliency and scanpath prediction on both MIT1003 and Salicon datasets. Besides the quantitative comparison, the qualitative results prove the effectiveness of our model. 

As future work, we plan to integrate the temporal dimension. We will also modify the loss function used to train the scanpath prediction branch in order to consider the saliency and the interdependence between both saliency and fixation points. Finally, we will test more advanced architectures instead of VGG.

{\small
\bibliographystyle{IEEEbib}
\bibliography{strings,refs}
}
\end{document}